\documentclass[runningheads]{llncs}

\usepackage{eccv}

\usepackage{eccvabbrv}

\usepackage{graphicx}
\usepackage{booktabs}
\usepackage{multirow}     
\usepackage{makecell}     
\usepackage[table]{xcolor}
\usepackage{soul}         
\usepackage{colortbl}     
\usepackage{subcaption} 
\usepackage{wrapfig}

\usepackage[accsupp]{axessibility}  

\usepackage[pagebackref,breaklinks,colorlinks,citecolor=eccvblue]{hyperref}

\usepackage{orcidlink}

\providecommand{\Letter}{\ensuremath{\star}}

\begin{document}

\title{MIMFlow: Integrating Masked Image Modeling with Normalizing Flows for End-to-End Image Generation}

\titlerunning{MIMFlow}

\author{Yang Chen\inst{1,2} \and
Xiaowei Xu\inst{2} \and
Shuai Wang\inst{1} \and
Xinwen Zhang\inst{1,2} \and
Qiushi Guo\inst{2} \and
Tiezheng Ge\inst{2} \and
Limin Wang\inst{1,3,\Letter}}

\authorrunning{Y.~Chen et al.}

\institute{State Key Laboratory for Novel Software Technology, Nanjing University \and
Alibaba Group \quad \textsuperscript{3}\enspace Shanghai AI Lab\\
\Letter~Corresponding author: lmwang@nju.edu.cn \\
Code: \url{https://github.com/MCG-NJU/MIMFlow}}

\maketitle

\begin{abstract}
Normalizing Flows (NFs) are powerful generative models capable of exact density estimation and sampling. However, their strict invertibility often forces the model to exhaust its capacity on low-level pixel details, hindering the capture of high-level semantic structures. While Masked Image Modeling (MIM) has excelled in representation learning, its integration into generative pipelines has remained largely modular and disjointed. In this paper, we propose MIMFlow, a unified end-to-end framework that jointly optimizes latent semantics, pixel reconstruction, and generative flow. By employing a VAE encoder to infer semantic latent from masked images, MIMFlow achieves a principled decoupling of the generative task: the Normalizing Flow focuses on modeling a simplified, low-frequency semantic manifold, while a specialized decoder handles high-frequency synthesis. This design effectively resolves the inherent capacity bottleneck of NFs, allowing the model to prioritize global structural coherence over redundant noise. Empirical results on ImageNet 256$\times$256 show that MIMFlow-L reaches 71.3\% linear probing accuracy and an FID of 2.50. Despite using only 128 tokens (50\% fewer than standard models), it yields a 32.8\% performance gain over similar-scale NF baselines.
\end{abstract}

\section{Introduction}
Normalizing Flows (NFs) are generative models that map a complex data distribution to a simple prior distribution through a series of invertible and differentiable transformations~\cite{realnvp, glow, zhai2024tarflow, flowback, simflow, farmer, tschannen2024jetformer, NF-CoT, NTM, starflow2, starflow-v, MIR_1}. A primary advantage of NFs is their ability to perform both exact density estimation and sampling within a single network. Besides, from the perspective of flow matching, NFs can be viewed as the expansion of an Ordinary Differential Equation (ODE), allowing for end-to-end optimization of the entire flow process~\cite{neuralode}. Recent work, such as SimFlow~\cite{simflow}, has further demonstrated that NFs can provide highly expressive priors for Variational Autoencoders (VAEs) by leveraging their likelihood estimation capabilities. However, strict invertibility forces NFs to prioritize low-level pixel details over high-level semantics, exhausting model capacity and hindering generative quality.

\label{sec:intro}
\begin{figure}[tbp] 
    \centering 
    \includegraphics[width=\textwidth]{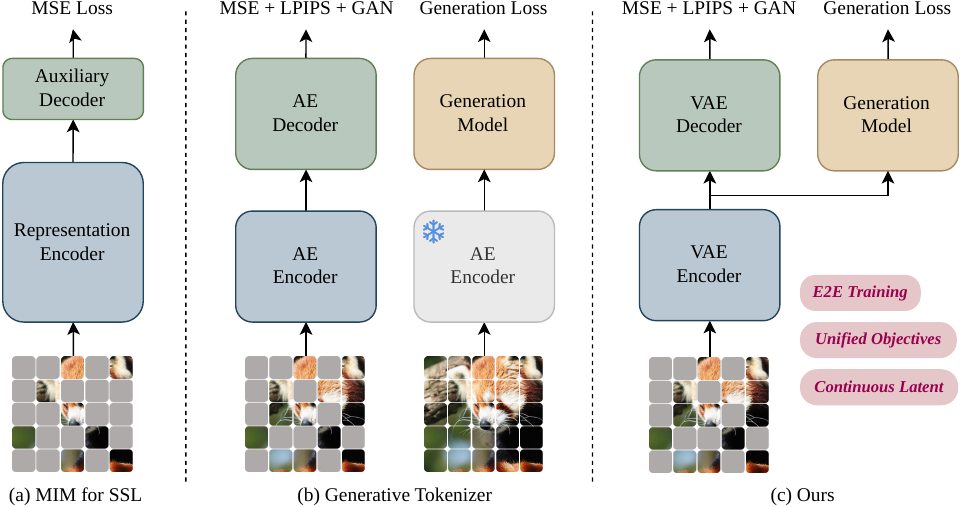} 
    \caption{\textbf{MIM in Different Paradigms.} (a) Self-Supervised Learning: Employs high-ratio masking as a self-supervised proxy task for representation learning. (b) Generative Tokenizers: A two-stage approach where the latent space is pre-trained with MIM before training a separate generative model. (c) MIMFlow (Ours): A unified framework that jointly optimizes latent semantics, pixel reconstruction, and generative flow in an end-to-end manner.} 
    \label{fig:intro} 
\end{figure}

While NFs struggle to capture global structures, Masked Image Modeling (MIM, ~\cref{fig:intro}a) has established itself as a cornerstone of self-supervised representation learning~\cite{MAE,simmim}. Despite its success in discriminative tasks, the role of MIM in generative modeling remains relatively under-explored. Recent literature suggests that integrating discriminative features can significantly bolster generative performance, with some approaches distilling knowledge from pre-trained representation models to guide the generation process~\cite{repa, lee2024repae, zheng2024lightningdit}. Furthermore, RAE~\cite{RAE, RAEv2} directly employs representation models as VAE encoders. Specifically, they found that MAE lags behind DINO in these generative frameworks. 
This likely stems from the mismatch between MAE’s high-mask reconstruction and the continuous distribution modeling required for synthesis. \textbf{Thus, whether semantic-focused MIM can effectively enhance generative models remains an open question.}

Recent attempts to incorporate MIM into generative pipelines have primarily focused on refining visual tokenizers. Specifically, works such as MAETok~\cite{MAETok} and DeTok~\cite{DeTok} have pioneered the use of masking as a denoising or discriminative objective to enhance the robustness of latent structures. Building on this, VTP~\cite{VTP} demonstrated that unifying MIM with contrastive learning can further foster a high-level understanding essential for tokenizer scalability. However, these methods largely treat the tokenizer as a modular, standalone precursor, separating representation learning from the core generative process (~\cref{fig:intro}b). In contrast, our method jointly optimizes representation, reconstruction and generation in an end-to-end framework (~\cref{fig:intro}c). This design enables a principled decoupling of the generative task: the NF is dedicated to modeling the low-frequency semantic manifold, while a specialized decoder handles high-frequency synthesis. By reducing the burden on NFs to capture pixel-level noise, MIMFlow mitigates a central capacity bottleneck of traditional NFs and offers an NF-oriented recipe for more semantic latent modeling.

Specifically, our MIMFlow employs a VAE encoder with learnable tokens to extract stable latent representations from masked images. These latents are then optimized through a dual-objective framework: a NF performs exact density estimation, while a decoder focuses on pixel-level reconstruction. This architecture facilitates a strategic division of labor: by mandating the encoder to infer missing spatial context, the resulting latent space is biased toward global structural coherence rather than redundant local noise. Consequently, the NF can model a simpler semantic manifold instead of dense pixel-level correlations. Crucially, linear probing evaluations reveal a substantial increase in classification accuracy within the latent space, validating the enhanced semantic quality of our representations. Finally, on the ImageNet 256×256 benchmark, MIMFlow improves similar-scale NF baselines while using fewer latent tokens.

In summary, our main contributions are as follows:
\begin{itemize}
    \item \textbf{MIMFlow Framework:} We propose an end-to-end framework that unifies representation and generation, moving beyond the modular limitations of existing tokenizer-based methods.
    \item \textbf{Principled Decoupling:} We introduce a strategy to decouple semantic modeling from pixel-level synthesis, effectively resolving the inherent capacity bottleneck of NFs.
    \item \textbf{Empirical Excellence:} We demonstrate that MIMFlow significantly enhances latent semantics and achieves superior generative performance on the ImageNet $256 \times 256$ benchmark.
\end{itemize}

\section{Related Work}
\textbf{Normalizing Flows}~\cite{vnf,papamakarios2021normalizing,kobyzev2020normalizing,dinh2014nice,realnvp,glow,draxler2024free,giaquinto2020gradient,draxler2024universality,MIR_2,MIR_3} provide a mathematically principled framework for bidirectional mapping between data and latent spaces. Early works like RealNVP~\cite{realnvp} and Glow~\cite{glow} established exact log-likelihood estimation through coupling layers, though they struggled to scale to high-resolution synthesis. Recent advancements~\cite{tschannen2024jetformer, starflow, zhai2024tarflow, flowback} have revitalized the field by integrating autoregressive Transformers and latent-space modeling, effectively leveraging NFs for both high-fidelity generation and semantic alignment. Building on this, SimFlow~\cite{simflow} demonstrates that NFs can serve as highly expressive probability estimators to replace restrictive VAE priors, significantly enhancing generative performance. While these models focus on architectural scaling or post-hoc alignment, our work is the first to integrate MIM objectives directly into the end-to-end training of NFs. By unifying self-supervised representation learning with generative flow optimization, we fully exploit the dual potential of NFs for simultaneous robust feature extraction and high-quality synthesis.


\noindent
\textbf{Masked Image Modeling} has emerged as a dominant paradigm in self-supervised learning, with methods like MAE~\cite{MAE} and SimMIM~\cite{simmim} demonstrating that reconstructing masked inputs forces encoders to learn robust structural features. While initially designed for discriminative tasks, it has also been utilized in generative frameworks. Specifically, recent literature explores distilling knowledge from pre-trained vision foundation models to guide diffusion processes~\cite{repa, zheng2024lightningdit}. Within the context of visual tokenizers, MAETok~\cite{MAETok} and DeTok~\cite{DeTok} employ masking as a denoising or discriminative auxiliary task to enhance latent robustness. Unlike methods above, MIMFlow integrates MIM directly into end-to-end generative training, ensuring that the generative flow is conditioned on highly compressed, semantic-rich features.

\noindent
\textbf{End-to-End Joint Training.} Generative models typically follow a two-stage pipeline—training a VAE for reconstruction and then a generative model on the frozen latent space. However, this can lead to a mismatch where high reconstruction fidelity fails to translate into generative quality. To bridge this gap, REPA-E~\cite{lee2024repae} and SimFlow~\cite{simflow} have pioneered end-to-end joint training, with SimFlow optimizing NFs and VAEs simultaneously from scratch. While our MIMFlow adopts this end-to-end philosophy, it distinguishes itself by incorporating a masked bottleneck to explicitly decouple semantic modeling from texture synthesis. This approach shields the NF from high-frequency noise and ensures the learned latent space is inherently optimized for global structure.

\section{Method}
\begin{figure}[tbp] 
    \centering 
    \includegraphics[width=\textwidth]{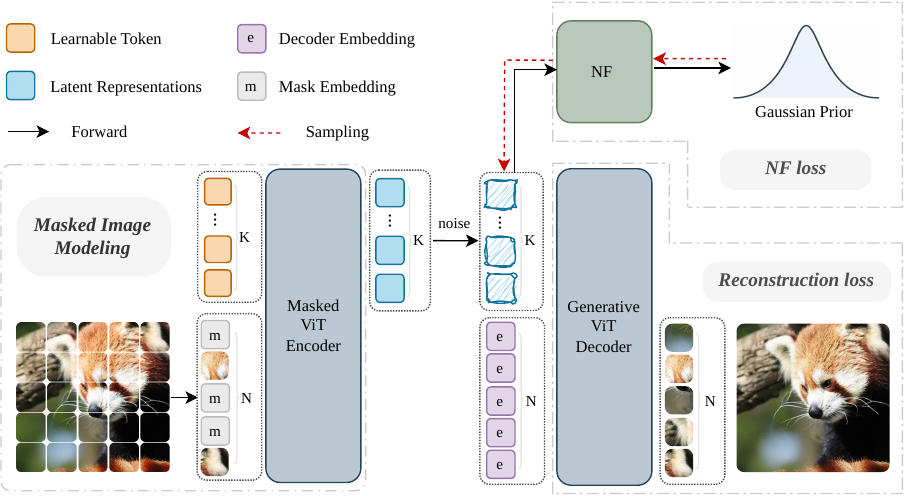} 
    \caption{\textbf{Structure of MIMFlow.} $N$ is the number of image patches, $K$ is the number of learnable latent query tokens, $\mathbf{m}$ is the binary mask, and $e$ denotes learnable decoder embeddings.} 
    \label{fig:main} 
\end{figure}
In this section, we present the architecture and mathematical formulation of \textbf{MIMFlow}, a unified generative framework that integrates MIM with Normalizing Flows. As illustrated in \cref{fig:main}, the framework comprises three core modules. First, a \textbf{Masked Encoder} ($E_\phi$) extracts robust latent representations by capturing global structural semantics from masked images. Second, a \textbf{Latent Normalizing Flow} ($f_\theta$) performs probabilistic modeling by mapping these representations to a Gaussian prior, enabling exact density estimation as the prior for VAE modeling. Finally, a \textbf{Generative Decoder} ($D_\psi$) handles pixel-level reconstruction and texture synthesis. This hierarchical design ensures that the flow model focuses on global structure while the decoder handles local details. In the following, we detail the latent extraction mechanism, the probabilistic modeling framework, and our multi-stage optimization strategy involving auxiliary supervision and adversarial refinement.

\subsection{Semantic Latent Extraction via Learnable Tokens}
\label{sec:latent_extraction}

A fundamental challenge in integrating MIM with NFs lies in the construction of a stable and expressive latent space. Conventional MIM backbones, such as MAE~\cite{MAE} and SimMIM~\cite{simmim}, present inherent difficulties for density estimation. MAE only processes visible patches, resulting in a latent sequence whose length and positional context vary with the random mask pattern, which imposes an intractable burden on the NF to learn a consistent distribution. Conversely, while SimMIM maintains a fixed sequence length by utilizing mask tokens, the information density within each token fluctuates significantly depending on its masking status. Such stochasticity prevents the NF from converging on a stable semantic manifold.

To resolve these issues, we introduce a \textbf{Learnable Token Bottleneck} to extract fixed-dimensional latent representations. We posit that since masked images contain lower information density than full images, they can be more efficiently compressed into a compact latent space. Formally, given an input image $\mathbf{x}$, we partition it into $N$ patches and apply a random masking ratio. The resulting $N$ tokens (including mask tokens) are concatenated with $K$ learnable query tokens, where $K < N$. These $N+K$ tokens are processed by a bidirectional Transformer encoder $E_\phi$, enabling the learnable queries to aggregate global semantic information from the partially observed image through self-attention.

After the encoding stage, we extract only the $K$ refined query tokens as our latent representation $\mathbf{z} \in \mathbb{R}^{K \times D}$. Following the practice of SimFlow~\cite{simflow}, we inject additive Gaussian noise with a fixed variance to $\mathbf{z}$ during training to facilitate continuous density modeling:
\begin{equation}
    \mathbf{\hat{z}} = \mathbf{z} + \sigma \epsilon, \quad \epsilon \sim \mathcal{N}(0, \mathbf{I}),
\end{equation}
where $\sigma$ is a hyperparameter. This $\mathbf{\hat{z}}$ serves a dual purpose within our framework: 
\begin{itemize}
    \item \textbf{Generative Modeling:} It is treated as a 1D sequence and fed into the NF $f_\theta$ for exact likelihood estimation, and reverse sampling.
    \item \textbf{Reconstruction:} It is passed to the decoder $D_\psi$, where it is combined with learnable image embeddings $e$ to reconstruct the original pixel-level signal through cross-modal attention.
\end{itemize}

This design provides several strategic advantages. First, the fixed-length $K$ tokens provide a stable target for the NF, regardless of the random mask positions. Second, the $K < N$ bottleneck forces the model to discard local pixel redundancies and concentrate on high-level structural semantics. Finally, the flexible masking ratio during training enhances the robustness of the latent space, ensuring that the NF models a manifold that is truly representative of the underlying global data structure.



\subsection{Probabilistic Modeling and Joint Optimization}

In this section, we formulate MIMFlow as a conditional generative model grounded in the Variational Inference (VI) framework. Specifically, we treat the encoder $E_\phi$ and decoder $D_\psi$ as a Variational Autoencoder (VAE) where the standard Gaussian prior is replaced by a high-capacity Normalizing Flow $f_\theta$. Unlike standard VAEs that take the complete image as input, our encoder is conditioned on the masked image $\tilde{\mathbf{x}} = \mathbf{x} \odot (1-\mathbf{m})$, where $\mathbf{m}$ denotes the binary mask.

\subsubsection{Variational Inference Perspective.}
Our objective is to maximize the log-likelihood of the data distribution $p(\mathbf{x}|\mathbf{m})$. Following the variational principle, we introduce an approximate posterior $q_\phi(\mathbf{z}|\tilde{\mathbf{x}})$ and derive the Evidence Lower Bound (ELBO):
\begin{equation}
    \log p(\mathbf{x}|\mathbf{m}) \ge \mathbb{E}_{q_\phi(\mathbf{z}|\tilde{\mathbf{x}})} [\log p_\psi(\mathbf{x}|\mathbf{z})] - D_{\text{KL}}(q_\phi(\mathbf{z}|\tilde{\mathbf{x}}) \parallel p_\theta(\mathbf{z})) = \mathcal{L}_{\text{ELBO}},
    \label{eq:elbo}
\end{equation}
where $p_\psi(\mathbf{x}|\mathbf{z})$ is the reconstruction likelihood and $p_\theta(\mathbf{z})$ is the prior distribution modeled by the Normalizing Flow. 

In our framework, we define $q_\phi(\mathbf{z}|\tilde{\mathbf{x}})$ as a Gaussian distribution with fixed variance $\sigma^2$ centered at the encoder's output $E_\phi(\tilde{\mathbf{x}})$. This simplifies the KL divergence term to the cross-entropy between the posterior and the flow-based prior (omitting constant entropy terms):
\begin{equation}
    D_{\text{KL}}(q_\phi(\mathbf{z}|\tilde{\mathbf{x}}) \parallel p_\theta(\mathbf{z})) \propto - \mathbb{E}_{q_\phi(\mathbf{z}|\tilde{\mathbf{x}})} [\log p_\theta(\mathbf{z})] + C.
\end{equation}

\subsubsection{Density Estimation via Normalizing Flow.} The prior $p_\theta(\mathbf{z})$ is parameterized by an invertible transformation $f_\theta$ that maps the complex latent $\mathbf{z}$ to a simple base distribution $\mathbf{\epsilon} \sim \mathcal{N}(0, \mathbf{I})$. Using the change-of-variables formula, the exact log-likelihood of the latent $\mathbf{z}$ can be computed as:
\begin{equation}
    \log p_\theta(\mathbf{z}) = \log p_{\epsilon}(f_\theta(\mathbf{z})) + \log \left| \det \frac{\partial f_\theta(\mathbf{z})}{\partial \mathbf{z}} \right|,
    \label{eq:flow_nll}
\end{equation}
where the first term represents the density under the Gaussian prior, and the second term is the log-determinant of the Jacobian, accounting for the volume change induced by the transformation. By optimizing this term, the NF effectively learns to warp the simple Gaussian into a sophisticated semantic manifold that captures the global dependencies of the image.

\subsubsection{Joint Training Objective.} Combining the reconstruction and the flow-based prior, we arrive at the total loss function for MIMFlow. The reconstruction term $\mathbb{E}_{q_\phi} [\log p_\psi(\mathbf{x}|\mathbf{z})]$ is implemented as a combination of $\ell_2$ loss and perceptual loss to ensure both pixel-level fidelity and semantic coherence:
\begin{equation}
    \mathcal{L}_{\text{rec}} = \|\mathbf{x} - D_\psi(\mathbf{z})\|_2^2 + \lambda_{\text{perc}} \mathcal{L}_{\text{LPIPS}}(\mathbf{x}, D_\psi(\mathbf{z})),
\end{equation}
where $\mathbf{z} \sim q_\phi(\mathbf{z}|\tilde{\mathbf{x}})$. Notably, since $\mathbf{z}$ is extracted from the masked image $\tilde{\mathbf{x}}$, the decoder is forced to perform both \textit{denoising} and \textit{inpainting}, which encourages the latent space to prioritize high-level structure over local noise.

The final joint loss is defined as:
\begin{equation}
    \mathcal{L}_{\text{prob}} = \mathcal{L}_{\text{rec}} + \beta \mathcal{L}_{\text{NF}},
    \label{eq:total_loss}
\end{equation}
where $\mathcal{L}_{\text{NF}} = - \log p_\theta(\mathbf{z})$ is the negative log-likelihood (NLL) provided by the NF. By jointly optimizing $\phi, \psi$ and $\theta$, MIMFlow ensures that the latent space is simultaneously optimized for representation, reconstruction and density estimation, leading to a more robust and expressive generative model.

\subsection{Auxiliary Supervision and Adversarial Refinement}
\label{sec:refinement}

To further enrich the latent representations and enhance the perceptual quality of the synthesized images, we incorporate auxiliary semantic supervision and a subsequent adversarial fine-tuning stage.

\subsubsection{Auxiliary Feature Prediction.} Following the design of generative tokenizers such as MAETok~\cite{MAETok}, we augment the training objective with an auxiliary discriminative task. In addition to the primary pixel decoder, a lightweight MLP-based auxiliary decoder $D_{aux}$ is employed to predict high-level features $\mathbf{F}_{target}$ (e.g., from DINO~\cite{oquab2023dinov2} or CLIP~\cite{clip}) directly from the latent $\mathbf{z}$. The auxiliary loss is defined as:
\begin{equation}
    \mathcal{L}_{\text{aux}} = \|D_{aux}(\mathbf{z}) - \text{sg}(\mathbf{F}_{target}(\mathbf{x}))\|_2^2.
\end{equation}
By supervising the latent space with pre-trained discriminative priors, this objective encourages the encoder to capture more robust semantic context beyond low-level pixel statistics. During the end-to-end training phase, the total loss is formulated as $\mathcal{L} = \mathcal{L}_{\text{rec}} + \beta \mathcal{L}_{\text{NF}} + \gamma \mathcal{L}_{\text{aux}}$, ensuring that the Normalizing Flow models a latent distribution that is both reconstructive and semantically rich.

\subsubsection{Adversarial Fine-tuning.} While the joint training phase establishes a structured latent space, the resulting reconstructions often lack the high-frequency details necessary for photorealistic generation. To address this, we perform a targeted fine-tuning of the pixel decoder $D_\psi$. In this stage, we introduce a patch-based discriminator $\mathcal{D}$ and optimize the model using a combination of reconstruction loss and GAN loss:
\begin{equation}
    \mathcal{L}_{\text{FT}} = \mathcal{L}_{\text{rec}} + \alpha \mathcal{L}_{\text{GAN}}(D_\psi, \mathcal{D}).
\end{equation}
Crucially, during fine-tuning, the encoder $E_\phi$ continues to receive the \textit{masked} image $\tilde{\mathbf{x}}$ as input. This preserves \textbf{distributional consistency} in the latent space: the decoder continues to observe latents from the same masked posterior family that the NF was trained to model, while sampling draws latents from the corresponding NF-modeled manifold rather than relying on an explicit mask. This allows the decoder to focus on synthesizing sharp textures without disrupting the probabilistic alignment between the generative flow and the latent representations.

\section{Experiment}
\label{sec:exp}
\subsection{Experimental Setup}

\subsubsection{Datasets and Evaluation Metrics}
We conduct our experiments on the ImageNet dataset at a $256 \times 256$ resolution~\cite{Deng2009ImageNet:Database}. To comprehensively evaluate the generative performance, we report the Fréchet Inception Distance (FID)~\cite{heusel2017gans}, Inception Score (IS)~\cite{salimans2016improved}, Precision, and Recall~\cite{kynkaanniemi2019improved}, all calculated using the evaluation suite provided by ADM \cite{dhariwal2021diffusion}. For reconstruction quality, we compute the reconstruction FID (rFID) on the ImageNet validation set. Furthermore, to assess the semantic quality of the learned latent space, we perform linear probing on the validation set following REPA's \cite{repa} protocol, reporting the top-1 classification accuracy.

\subsubsection{Implementation Details}
Our framework consists of two primary components: the Normalizing Flow prior and the Transformer-based Autoencoder.

\begin{wraptable}{r}{0.5\textwidth} 
    \centering
    \vspace{-35pt} 
    \small
    \setlength{\tabcolsep}{0.3mm}
    \caption{\textbf{Baseline Construction} (last line), evaluated with 10K samples.}
    \label{tab:baseline_evolution}
    \begin{tabular}{lccc}
        \toprule
        & w/o CFG &\multicolumn{2}{c}{w/ CFG} \\ 
        \cmidrule(lr){2-4}
        Configuration & gFID $\downarrow$ & gFID $\downarrow$ & sFID $\downarrow$ \\ 
        \midrule
        STARFlow-L & 50.22 & 7.79 & 20.54 \\
        \quad $-$ Softplus & 48.24 & 7.65 & 18.98 \\
        \quad $+$ Gated Attn. & 47.44 & 7.45 & 18.64 \\
        \rowcolor[gray]{0.95} 
        \quad $+$ End-to-End & \textbf{11.24} & \textbf{5.99} & -- \\ 
        \bottomrule
    \end{tabular}
    \vspace{-20pt} 
\end{wraptable}

\begin{figure}[t]
    \centering
    \includegraphics[width=\linewidth]{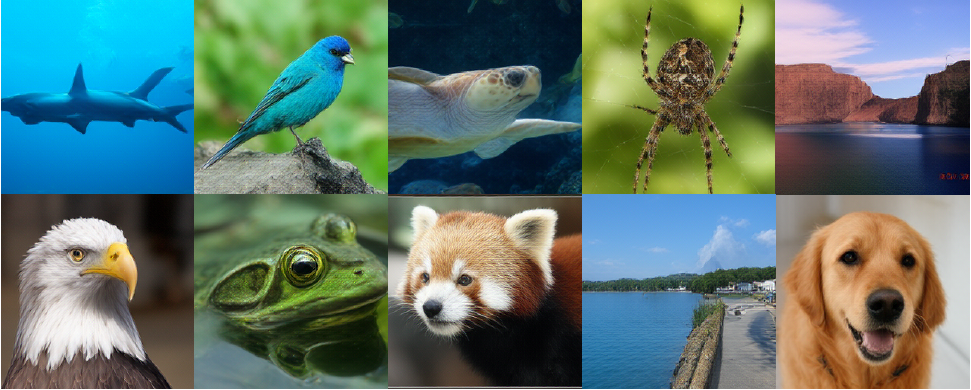}
    \caption{\textbf{Selected Samples} on ImageNet 256 × 256 from MIMFlow-L. We use classifier-free guidance equal to 2.0.}
    \label{fig:visualization}
\end{figure}
\paragraph{Normalizing Flow Architecture} We adopt the STARFlow~\cite{starflow} architecture for density estimation in the latent space. Our model, designated as \textbf{STARFlow-L}, comprises approximately 482M parameters with a hidden dimension of 1024. The backbone consists of seven 2-layer blocks followed by a final 20-layer block. 

We iteratively construct our baseline starting from a STARFlow-L model trained on a fixed VAE latent space. The refinement process involves: (1) removing the \textbf{softplus} operation on the scaling factors to improve numerical flexibility; (2) incorporating \textbf{gated attention}~\cite{gatedattn} mechanisms to enhance feature interaction; and (3) transitioning to full \textbf{end-to-end (e2e)} training with GAN loss, and without auxiliary loss. Relative to SimFlow-L, this baseline changes only the NF backbone, which makes the close FID (3.70 vs.\ 3.72) an apples-to-apples check. The performance gains from each stage are summarized in ~\cref{tab:baseline_evolution}. For optimization, we use the AdamW optimizer with a learning rate of $1 \times 10^{-4}$ and a global batch size of 256. The final model is trained end-to-end for 90 epochs, followed by 2 additional epochs of decoder fine-tuning.

\begin{wraptable}{r}{0.4\textwidth} 
    \centering
    \vspace{-35pt} 
    \caption{Comparison of Architectural Complexity}
    \vspace{8pt}
    \label{tab:arch_comparison}
    \begin{tabular}{lcc}
    \toprule
    Model & Parameters & FLOPs \\ \hline
    SD-VAE & 83.7M & 446.21G \\
    MAETok & 172.0M & 71.25G \\ \bottomrule
    \end{tabular}
    \vspace{-10pt} 
\end{wraptable}
\paragraph{Transformer-based Autoencoder} To accommodate the requirements of MIM, we replace the traditional convolutional stages of the SD-VAE\footnote{\url{https://huggingface.co/stabilityai/sd-vae-ft-ema}}~\cite{rombach2022high} with Transformer-based MAETok~\cite{MAETok}. Notably, we train the entire framework from scratch without loading any pre-trained parameters for the MAETok backbone.
Our latent space is configured with 128 tokens, each having a dimensionality of 64, and is integrated with 1D positional encodings. During training, we adopt a random masking strategy to facilitate the MIM objective, with a default mask ratio ranging from 0.4 to 0.6. As shown in ~\cref{tab:arch_comparison}, while the ViT-B backbone increases the parameter count, it significantly reduces the computational overhead in terms of FLOPs compared to the convolutional VAE, making it more suitable for high-resolution processing


\subsection{Main Results}
\begin{table}[t]
\scriptsize 
    \centering
    \caption{\textbf{System performance comparison} on ImageNet $256\times256$ class-conditioned generation. {\setlength{\fboxsep}{1.5pt}\colorbox[gray]{0.9}{Gray}} rows denote larger-scale models within the latent normalizing flow category, provided for reference.}
    \setlength{\tabcolsep}{0.7mm} 
\begin{tabular}{>{\raggedright\arraybackslash}p{2.6cm} cc cccc cccc} 
\toprule
\multirow{2}{*}{Method} & \multirow{2}{*}{\makecell{\#Tokens}} & \multirow{2}{*}{\#Params} & \multicolumn{4}{c}{W/o guidance} & \multicolumn{4}{c}{W/ guidance} \\
\cmidrule(lr){4-7} \cmidrule(lr){8-11}
 & & & gFID$\downarrow$ & IS$\uparrow$ & Prec.$\uparrow$ & Rec.$\uparrow$ & gFID$\downarrow$ & IS$\uparrow$ & Prec.$\uparrow$ & Rec.$\uparrow$ \\
\midrule
\multicolumn{11}{l}{\textit{Pixel Space}} \\
\arrayrulecolor{black!30}\midrule
ADM~\cite{dhariwal2021diffusion} &  -  &  554M & 10.94 &  101.0 & 0.69 & 0.63 & 3.94 & 215.8 & 0.83 & 0.53\\
RIN~\cite{RIN} &  -  &  410M & 3.42  & 182.0  &  -   &  -   &  -   &   -    &  -   &  -  \\
PixelFlow~\cite{pixelflow} & 4096 & 677M & - & -   &   -  &  -   & 1.98 & 282.1 & 0.81 & 0.60 \\
PixNerd~\cite{pixnerd} & 1024 & 700M & -   &  -       &  - &  -  &  2.15 & 297.0 & 0.79 & 0.59 \\
SiD2~\cite{SID} &  -   &   -   & -   &  - &  -   &  -   &  1.38   &   -    &  -   &  - \\
TARFlow~\cite{zhai2024tarflow} & 1024 & 1.4B & - & - & - & - & 4.69 & - & - & - \\
JetFormer~\cite{tschannen2024jetformer} & 256 & 2.8B & - & - & - & -  & 6.64 & - & 0.69 & 0.56 \\
FARMER~\cite{farmer} & 1024 & 1.9B & - & - & - & - & 3.60 & 269.2 & 0.81 & 0.51\\
\arrayrulecolor{black}\midrule
\multicolumn{11}{l}{\textit{Latent Autoregressive}} \\
\arrayrulecolor{black!30}\midrule
VAR~\cite{var} &  680  &  2.0B  & 1.92 & 323.1  & 0.82 & 0.59 & 1.73 & 350.2 & 0.82 & 0.60\\
MAR~\cite{mar} &  256  &  943M  & 2.35 &  227.8 & 0.79 & 0.62 & 1.55 & 303.7 & 0.81 & 0.62\\
xAR~\cite{xar} &  -  & 1.1B  & - & - & - & - & 1.24 & 301.6 & 0.83 & 0.64\\
\arrayrulecolor{black}\midrule
\multicolumn{11}{l}{\textit{Latent Diffusion}} \\
\arrayrulecolor{black!30}\midrule
DiT~\cite{DiT} & 256 & 675M & 9.62 & 121.5 & 0.67 & 0.67 & 2.27 & 278.2 & 0.83 & 0.57 \\
MaskDiT~\cite{MaskDiT} & - & 675M & 5.69 & 177.9 & 0.74 & 0.60 & 2.28 & 276.6 & 0.80 & 0.61 \\
SiT~\cite{sit} & 256 & 675M & 8.61 & 131.7 & 0.68 & 0.67 & 2.06 & 270.3 & 0.82 & 0.59 \\
MDTv2~\cite{mdtv2} & 256 & 675M & - & - & - & - & 1.58 & 314.7 & 0.79 & 0.65 \\
REPA~\cite{repa} & 256 & 675M & 5.78 & 158.3 & 0.70 & 0.68 & 1.29 & 306.3 & 0.79 & 0.64 \\
VA-VAE~\cite{vavae} & 256 & 675M & 2.17 & 205.6 & 0.77 & 0.65 & 1.35 & 295.3 & 0.79 & 0.65 \\
DDT~\cite{DDT,uniDDT} & 256 & 675M & 6.27 & 154.7 & 0.68 & 0.69 & 1.26 & 310.6 & 0.79 & 0.65\\
REPA-E~\cite{lee2024repae} & 256 & 675M & 1.69 & 219.3 & 0.77 & 0.67 & 1.12 & 302.9 & 0.79 & 0.66 \\
RAE~\cite{RAE} & 256 & 839M & 1.51 & 242.9 & 0.79 & 0.63 & 1.13 & 262.6 & 0.78 & 0.67 \\
 \arrayrulecolor{black}\midrule
\multicolumn{11}{l}{\textit{Latent Normalizing Flows}} \\
\arrayrulecolor{black!30}\midrule
\rowcolor[gray]{0.9} FlowBack-XL~\cite{flowback} & 1024 & 831M & - & - & - & - & 4.18 & 240.8 & - & - \\
\rowcolor[gray]{0.9} STARFlow-XXL~\cite{starflow} & 1024 & 1.4B & - & - & - & - & 2.40 & - & - & - \\
\rowcolor[gray]{0.9} FAE-NF-XXL~\cite{FAE} & 256 & 1.4B & - & - & - & - & 2.67 & - & - & - \\
\rowcolor[gray]{0.9} SimFlow-XXL~\cite{simflow} & 256 & 1.4B & 10.13 & 124.7 & 0.71 & 0.61 & 1.91 & 284.4 & 0.82 & 0.60 \\
SimFlow-L~\cite{simflow} & 256 & 475M & 33.53 & - & - & - & 3.72 & - & -& - \\
Baseline (Ours) & 256 & 482M &  - & - & - & - & 3.70 & - & - & - \\
MIMFlow-L (Ours) & 128 & 482M & 3.64 & 158.6 & 0.78 & 0.60 & 2.50 & 233.5 & 0.82 & 0.57 \\
\arrayrulecolor{black}\bottomrule
\end{tabular}
\label{tab:imagenet256_sota}
\end{table}

We compare MIMFlow-L with state-of-the-art (SOTA) generative models on the ImageNet $256 \times 256$ benchmark, including Pixel-space models, Latent Autoregressive (AR) models, Latent Diffusion Models (LDMs), and existing Latent Normalizing Flows. The quantitative results are summarized in Table~\ref{tab:imagenet256_sota}. Besides, qualitative results are shown in Fig.~\ref{fig:visualization}, where MIMFlow-L produces high-fidelity images with consistent global structures.

\noindent
\textbf{Superiority within the NF Paradigm.} 
MIMFlow-L demonstrates a significant performance leap over existing Normalizing Flow baselines. Compared to the closely related SimFlow-L, which shares a similar parameter count ($\sim$480M), MIMFlow-L improves the FID from 3.72 to \textbf{2.50}—a 32.8\% reduction. Notably, our model with only 482M parameters outperforms much larger models like FAE-NF-XXL (FID 2.67), which utilizes nearly 3 $\times$ the parameters (1.4B) and is built upon the advanced RAE latent space. It also closely approaches the performance of STARFlow-XXL (FID 2.40). This suggests that integrating MIM allows the flow model to learn a more efficient and structured semantic manifold, extracting higher generative value from the same parameter budget.

\noindent
\textbf{Efficiency via Token Compression.} 
A key highlight of MIMFlow is its token efficiency. While most latent models (e.g., DiT, LDM, SimFlow) operate on a $16 \times 16 = 256$ token grid, and some even require 1024 tokens (STARFlow, FlowBack), MIMFlow-L achieves strong NF results using only \textbf{128 tokens} for sampling.
\begin{enumerate}
\item\textit{Reduced Computational Complexity:} By halving the sequence length compared to standard VAE-based models, the computational overhead of the Normalizing Flow (which typically scales quadratically or involves deep Transformer blocks) is significantly reduced. 
\item\textit{Information Density:} This 128-token bottleneck supports our hypothesis in ~\cref{sec:latent_extraction}: because the NF is less exposed to high-frequency noise through MIM, it can represent the global image structure more compactly. This allows MIMFlow to maintain high precision (0.82) while operating at a fraction of the sequence length used by many competitors. Appendix Table~\ref{tab:performance_metrics} further reports the latency benefit of reducing the token count.
\end{enumerate}

\begin{table}[t]
\centering
\small
\setlength{\tabcolsep}{5pt}
\caption{\textbf{Hardware efficiency comparison} on ImageNet $256 \times 256$ using 8$\times$H800 GPUs. We use no auxiliary supervision in this comparison; the per-GPU batch sizes are 16 for training and 256 for inference.}
\label{tab:hardware_efficiency}
\begin{tabular}{lccccc}
\toprule
Model & Tokens & Params & Train Mem. & Train Speed & Sample / img \\
\midrule
SimFlow-L & 256 & 475M & 52.3GB & 2.83 it/s & 0.020s \\
\rowcolor[gray]{0.95}\textbf{MIMFlow-L} & \textbf{128} & \textbf{482M} & \textbf{37.6GB} & \textbf{3.11 it/s} & \textbf{0.011s} \\
\bottomrule
\end{tabular}
\end{table}

\noindent
We further compare L-scale flow models under the same hardware setting in \cref{tab:hardware_efficiency}. Despite using slightly more parameters (482M vs. 475M), MIMFlow-L reduces training memory from 52.3GB to 37.6GB, improves throughput from 2.83 to 3.11 iterations per second, and reduces per-image sampling time from 0.020s to 0.011s. This corresponds to a 28\% memory reduction, a 10\% throughput improvement, and nearly halved sampling time; the lower memory footprint also makes training feasible on 8$\times$A100 GPUs with a total batch size of 256.

\noindent
\textbf{Impact of Guidance and Semantic Quality.} 
As shown in ~\cref{tab:imagenet256_sota}, the gap between ``w/o guidance'' and ``w/ guidance'' performance is notably narrower for MIMFlow compared to early flow models. Without guidance, MIMFlow-L achieves an FID of 3.64, which is significantly better than the 10.13 reported for SimFlow-XXL. This indicates that the latent space learned through joint MIM and NF optimization is inherently more structured and semantically coherent, requiring less external guidance to produce high-quality samples. 

\noindent
\textbf{Comparison with Diffusion and AR Models.} 
While Latent Diffusion Models (like DiT and REPA) currently lead in FID, MIMFlow-L narrows the gap between ODE-based Normalizing Flows and these heavy-duty generative frameworks. Unlike diffusion models that require hundreds of denoising steps or complex scheduling, our flow-based approach offers exact density estimation and a deterministic mapping in a single continuous ODE trajectory. The competitive Precision (0.82) of MIMFlow-L underscores its ability to generate high-fidelity samples that faithfully respect the learned data manifold.

In summary, the results demonstrate that by decoupling semantic modeling from texture synthesis, MIMFlow-L mitigates a capacity bottleneck in traditional NFs and provides an efficient path for improving NF image generation.

\subsection{Ablation Study}
\begin{table}[tbp]
\centering
\footnotesize 
\caption{\textbf{Ablation studies} on ImageNet $256 \times 256$.
\textbf{Bold} indicates the default configuration. \textbf{Acc.} denotes linear probing accuracy on the encoder's representations \textit{before} the projection layer. \textbf{Mix} in ~\cref{tab:ablation_mask} represents a 50/50 probability of using \textit{None} and \textit{0.4--0.6} masking ratios. 10K images are sampled for evaluation.}
\label{tab:ablations_combined}
\begin{minipage}[t]{0.507\textwidth}
    \centering
    \subcaption{Masking Strategies}
    \label{tab:ablation_mask}
    \setlength{\tabcolsep}{0.3mm} 
    \resizebox{\textwidth}{!}{ 
    \begin{tabular}{lcccccc}
    \toprule
    Ratio & rFID$\downarrow$ & gFID$\downarrow$ & IS$\uparrow$ & Prec.$\uparrow$ & Rec.$\uparrow$ & Acc.$\uparrow$ \\
    \midrule
    None & 23.5 & 29.0 & 65.5 & 0.59 & 0.56 & 56.6 \\
    0.2--0.4 & 8.66 & 24.47 & 60.1 & 0.59 & 0.57 & 54.2 \\
    \rowcolor[gray]{0.95} \textbf{0.4--0.6} & \textbf{3.40} & \textbf{12.82} & \textbf{88.6} & \textbf{0.70} & \textbf{0.66} & \textbf{71.3} \\
    0.6--0.8 & 5.26 & 15.92 & 79.8 & 0.65 & 0.65 & 65.9 \\
    Mix & 6.58 & 26.98 & 51.9 & 0.58 & 0.55 & 45.7 \\
    \bottomrule
    \end{tabular}
    }
\end{minipage}
\hfill
\begin{minipage}[t]{0.483\textwidth}
    \centering
    \subcaption{Auxiliary Loss Targets}
    \label{tab:ablation_aux}
    \setlength{\tabcolsep}{0.5mm}
    \resizebox{\textwidth}{!}{
    \begin{tabular}{lccccc}
    \toprule
    Target & rFID$\downarrow$ & gFID$\downarrow$ & IS$\uparrow$ & Prec.$\uparrow$ & Rec.$\uparrow$ \\
    \midrule
    DINO & 4.00 & 12.89 & 91.0 & 0.70 & 0.66 \\
    D+HOG & \multicolumn{5}{c}{\textit{Training Collapsed}} \\
    \rowcolor[gray]{0.95} \textbf{D+CLIP} & \textbf{3.60} & \textbf{12.46} & \textbf{92.3} & \textbf{0.70} & \textbf{0.66} \\
    C+HOG & 4.18 & 20.77 & 56.0 & 0.63 & 0.59 \\
    All & 3.64 & 12.82 & 88.6 & 0.70 & 0.66 \\
    \bottomrule
    \end{tabular}
    }
\end{minipage}


\begin{minipage}{0.49\textwidth}
    \centering
    \subcaption{Token Number ($K$)}
    \label{tab:ablation_token}
    \setlength{\tabcolsep}{1.2mm}
    \resizebox{\textwidth}{!}{
    \begin{tabular}{lccccc}
    \toprule
    $K$ & rFID$\downarrow$ & gFID$\downarrow$ & IS$\uparrow$ & Prec.$\uparrow$ & Rec.$\uparrow$ \\
    \midrule
    64 & 5.61 & 12.78 & 91.8 & \textbf{0.71} & 0.65 \\
    \rowcolor[gray]{0.95} \textbf{128} & \textbf{3.60} & \textbf{12.46} & \textbf{92.3} & 0.70 & \textbf{0.66} \\
    192 & 24.59 & 30.42 & 68.0 & 0.57 & 0.57 \\
    \bottomrule
    \end{tabular}
    }
\end{minipage}
\hfill
\begin{minipage}{0.49\textwidth}
    \centering
    \subcaption{Latent Noise Scale ($\sigma$)}
    \label{tab:ablation_std}
    \setlength{\tabcolsep}{1.2mm}
    \resizebox{\textwidth}{!}{
    \begin{tabular}{lccccc}
    \toprule
    $\sigma$ & rFID$\downarrow$ & gFID$\downarrow$ & IS$\uparrow$ & Prec.$\uparrow$ & Rec.$\uparrow$ \\
    \midrule
    0.2 & 3.93 & 14.30 & 81.5 & 0.66 & 0.66 \\
    \rowcolor[gray]{0.95} \textbf{0.3} & \textbf{3.40} & \textbf{12.82} & \textbf{88.6} & \textbf{0.70} & \textbf{0.66} \\
    0.5 & 6.95 & 17.69 & 75.6 & 0.65 & 0.62 \\
    \bottomrule
    \end{tabular}
    }
\end{minipage}
\end{table}
To validate the effectiveness of the proposed components in MIMFlow, we conduct a series of ablation experiments on the ImageNet $256 \times 256$ benchmark. To ensure a fair comparison, all models in this section are trained end-to-end for 50 epochs. Unless otherwise specified, the reported generative metrics are calculated using 10K samples (50K in ~\cref{tab:imagenet256_sota}).

\noindent \textbf{The Necessity of Masked Decoupling.} 
~\Cref{tab:ablation_mask} demonstrates the critical role of the masking strategy under the same end-to-end training setup. Without masking (\textit{nomask}), the model exhibits the poorest performance (gFID 29.0), and its latent semantic quality (Acc. 56.6\%) is significantly lower than that of masked versions. This supports our hypothesis that unmasked training leaves the Normalizing Flow exposed to redundant pixel-level details. A mask ratio of 0.4--0.6 achieves the best balance, significantly enhancing both generative quality (gFID 12.82) and semantic representation (Acc. 71.3\%). Interestingly, a mixed strategy or low mask ratios lead to performance degradation, suggesting that a consistent and substantial information bottleneck is important for stabilizing the semantic manifold.

\begin{figure}[tbp]
    \centering
    \begin{subfigure}[b]{0.4\textwidth}
        \centering
        \includegraphics[width=\textwidth]{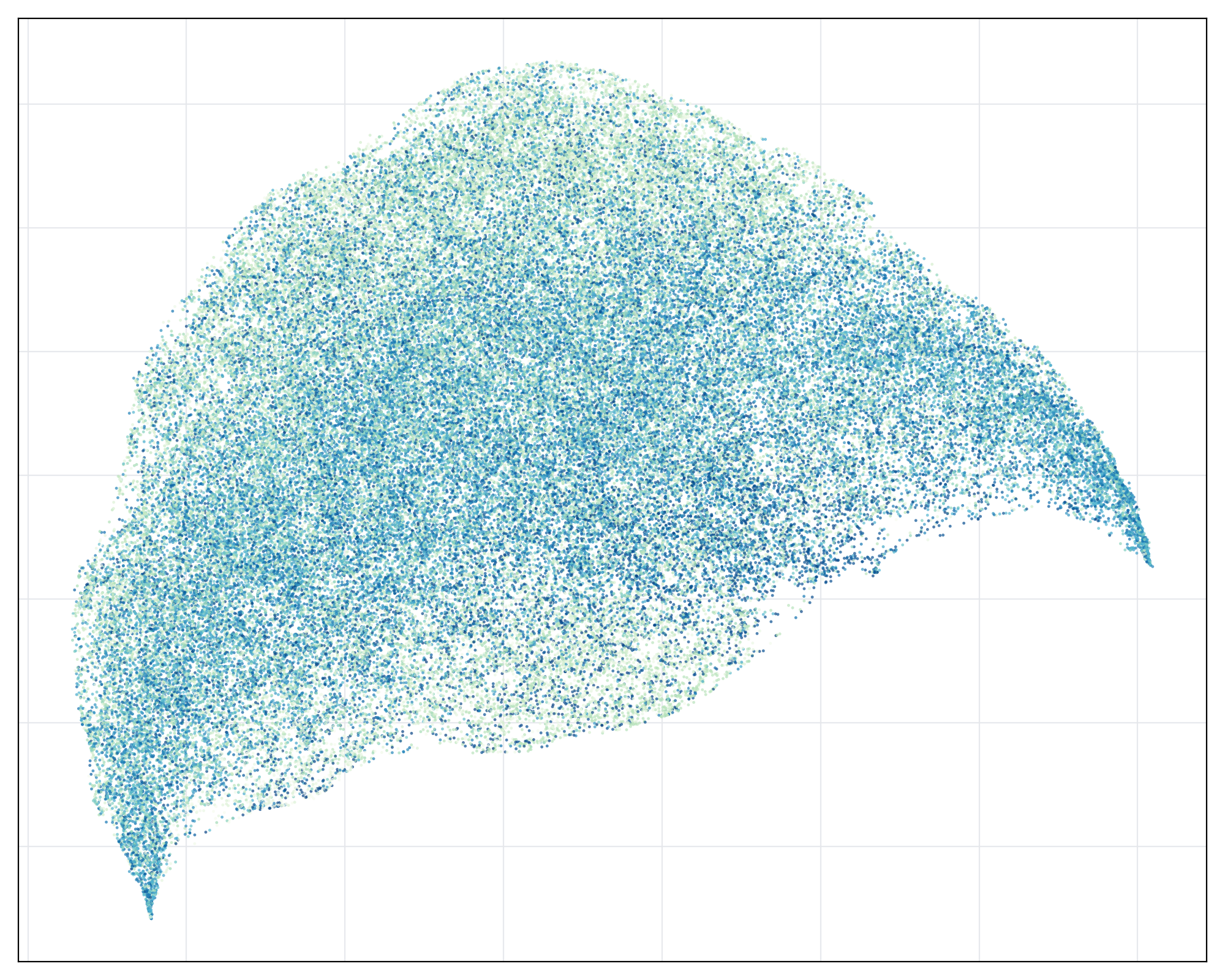} 
        \caption{SD-VAE}
        \label{fig:image2}
    \end{subfigure}
    \begin{subfigure}[b]{0.4\textwidth}
        \centering
        \includegraphics[width=\textwidth]{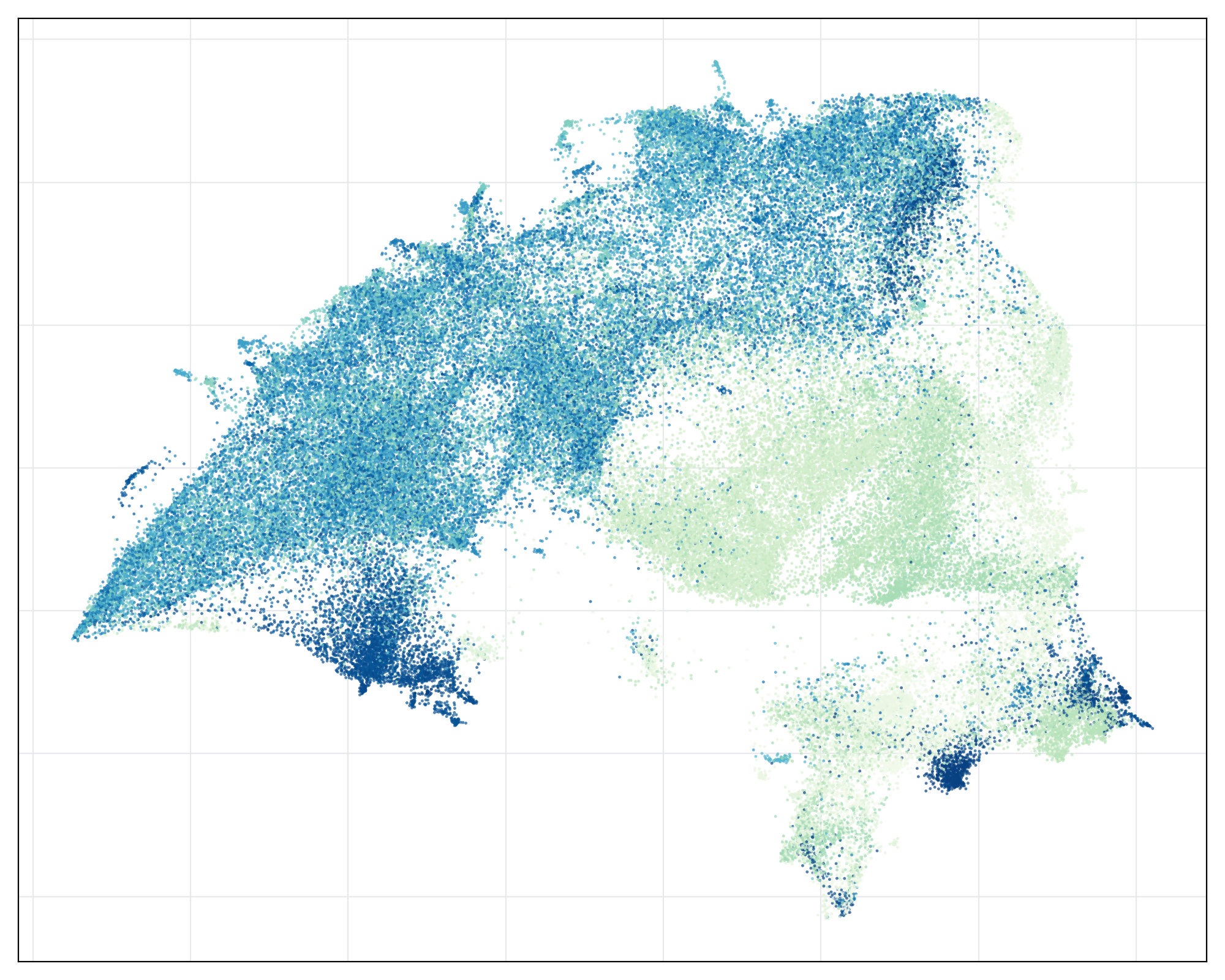} 
        \caption{MIMFlow}
        \label{fig:image1}
    \end{subfigure}
    \hfill 
    
    \caption{\textbf{UMAP visualization} on ImageNet of the learned latent
space from (a) SD-VAE; (b) MIMFlow. Colors indicate different classes. MIMFlow presents a more discriminative latent space.}
    \label{fig:umap}
\end{figure}
\noindent \textbf{Optimal Bottleneck Size.} 
As shown in ~\cref{tab:ablation_token}, the number of latent tokens $K$ serves as a physical bottleneck for information compression. While 64 tokens provide reasonable performance, increasing to 128 tokens yields the best reconstruction (rFID 3.60) and generation results. This observation aligns with the information density provided by our masking strategy: since a $256 \times 256$ image is partitioned into $16 \times 16 = 256$ patches and approximately 50\% of the patches are masked during training, the remaining information can be optimally represented by a compressed latent space of 128 tokens. However, expanding to 192 tokens leads to a sharp performance drop (gFID 30.42). This indicates that an excessively large latent space allows high-frequency noise to leak into the flow model, thereby violating the principled decoupling and hindering the NF's ability to model global structure.

\noindent \textbf{Synergy of Auxiliary Semantic Priors.} 
We investigate various auxiliary supervision signals (DINO, CLIP, HOG) in ~\cref{tab:ablation_aux}. The combination of DINO and CLIP features achieves the superior performance, as they provide complementary structural and semantic guidance. In contrast, incorporating low-level features like HOG either leads to training collapse or suboptimal results. This validates that the MIMFlow latent space is inherently optimized for high-level semantic abstractions rather than local gradient textures.

\noindent \textbf{Impact of Latent Stochasticity.} 
The additive Gaussian noise $\sigma$ is a crucial trick for NF training. As shown in ~\cref{tab:ablation_std}, $\sigma=0.3$ is the optimal value. Smaller values ($\sigma=0.2$) fail to sufficiently smoothen the manifold, while larger values ($\sigma=0.5$) introduce excessive variance that compromises reconstruction fidelity (rFID 6.95), confirming that a precise calibration of latent stochasticity is vital for end-to-end flow modeling.

\subsection{Analysis of Latent Space and Flow Dynamics}
\label{sec:analysis}
    

In this section, we further investigate why the proposed MIMFlow framework facilitates more effective generative modeling by analyzing the properties of the learned latent space and the flow transformations.

\noindent
\textbf{Semantic Discriminability.}
A core motivation of MIMFlow is to decouple high-frequency pixel noise from global structural semantics. We visualize the latent space $\mathbf{z}$ using UMAP projection in \cref{fig:umap}. Compared to the relatively entangled latent space of a standard SD-VAE, the MIMFlow latent space exhibits significantly clearer clustering corresponding to ImageNet categories. This enhanced discriminative power is quantitatively supported by the linear probing results in \cref{tab:ablation_mask}: our masked approach achieves a classification accuracy of \textbf{71.3\%}, a substantial improvement over the \textbf{56.6\%} of the unmasked baseline. These results confirm that the masking bottleneck effectively forces the latent manifold to prioritize high-level semantic coherence over local redundancies.
\begin{figure}[t]
    \centering
    \includegraphics[width=\linewidth]{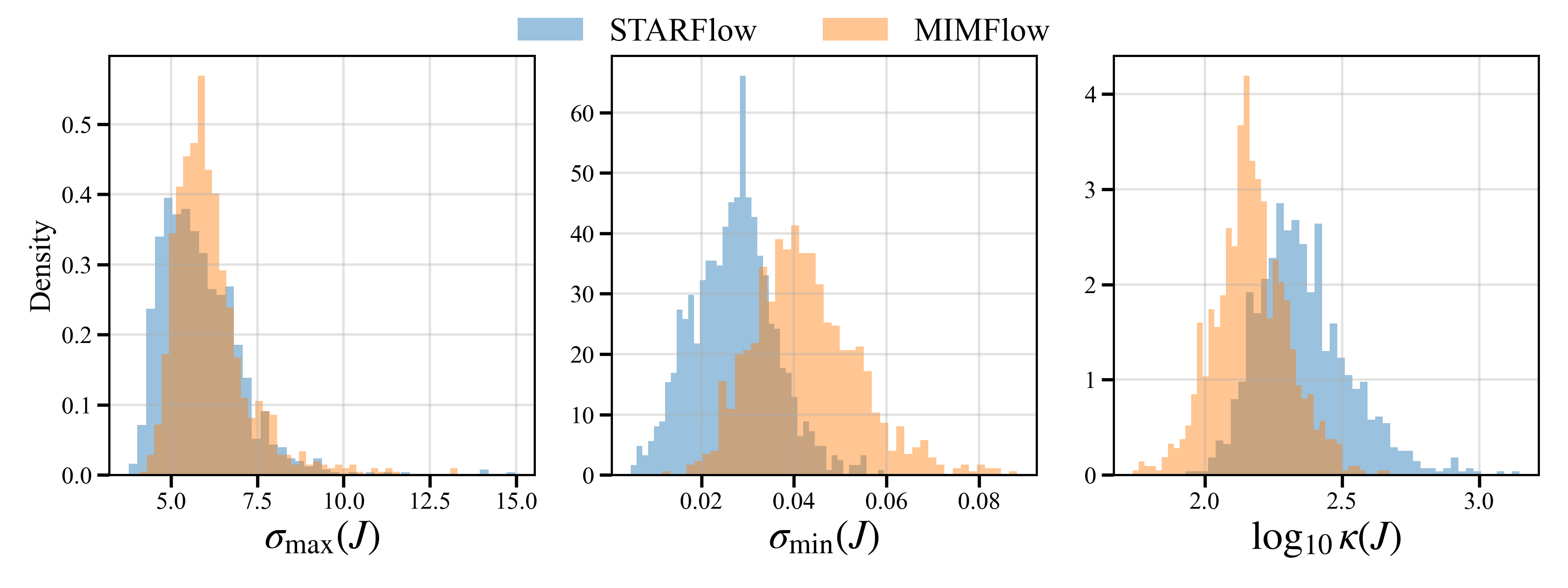}
    \caption{\textbf{Jacobian Spectral Analysis} of STARFlow and MIMFlow. The three panels report, from left to right, the empirical distributions of the largest singular value $\sigma_{\max}(J)$, the smallest singular value $\sigma_{\min}(J)$, and the log-condition number $\log_{10} \kappa(J)$ (with $\kappa(J) = \sigma_{\max}(J) / \sigma_{\min}(J)$).}
    \label{fig:jacobi}
\end{figure}

\noindent
\textbf{Jacobian Spectral Analysis.}
To assess the numerical stability and bijectivity of the learned Normalizing Flow, we analyze the spectral properties of the Jacobian $J = \partial f_\theta / \partial \mathbf{z}$ for both STARFlow and MIMFlow (\cref{fig:jacobi}). The analysis reveals several key insights:
\begin{enumerate}
    \item \textit{Enhanced Bijectivity:} MIMFlow exhibits a larger and more stable minimum singular value ($\sigma_{\min}$), keeping the Jacobian further from singularity and reducing the risk of ill-conditioned mappings.
    \item \textit{Superior Numerical Stability:} Compared to STARFlow, MIMFlow achieves a lower and more concentrated log-condition number. A lower condition number indicates a more well-conditioned transformation that is less sensitive to numerical perturbations. 
\end{enumerate}
These spectral properties suggest that by alleviating the burden on the NF to capture pixel-level details, MIMFlow learns a smoother flow field with less extreme spatial warping, which is consistent with more stable optimization and higher generative fidelity.

\section{Limitation}
Our experiments validate MIM-style masked semantic bottlenecks for latent NF models, but do not imply a general claim about MIM across all generative model families, such as diffusion or autoregressive models. This work also focuses on class-to-image generation on ImageNet; extending it to text-to-image generation, where prompt alignment and compositional semantics are central, remains future work. Finally, while we use standard ImageNet generation metrics such as FID, IS, precision/recall, and linear probing, newer metrics such as Representation Fréchet Loss~\cite{fdloss} may provide complementary signals and are worth exploring.

\section{Conclusion}
In this paper, we presented MIMFlow, a unified end-to-end generative framework that integrates Masked Image Modeling with Normalizing Flows. By introducing a masked bottleneck with learnable tokens, we achieve a principled decoupling of generative tasks: the Normalizing Flow focuses on modeling the low-frequency semantic manifold, while a specialized decoder handles high-frequency texture synthesis. This design mitigates the capacity bottleneck of traditional NFs, which often spend representational power on redundant pixel-level noise.
Empirical results on ImageNet $256\times256$ demonstrate that MIMFlow-L improves similar-scale NF baselines, achieving an FID of 2.50. Notably, our model achieves this with only 128 tokens--a 50\% reduction compared to standard latent generative models--while maintaining high semantic discriminability, as evidenced by a 71.3\% linear probing accuracy. Our analysis of the Jacobian spectrum further suggests that this decoupling leads to better-conditioned flow transformations. Ultimately, MIMFlow provides a new perspective on unifying self-supervised representation learning and probabilistic modeling, paving the way for more efficient and semantically-aware generative systems.

\section*{Acknowledgements}
This work is supported by the Basic Research Program of Jiangsu (No. BK20250009), the Fundamental Research Funds for the Central Universities (No.020214380140), the Fundamental and Interdisciplinary Disciplines Breakthrough Plan of the Ministry of Education of China (No. JYB2025XDXM118), the Collaborative Innovation Center of Novel Software Technology and Industrialization, Alibaba Group through Alibaba Innovative Research Program.
%
%
\appendix
\renewcommand{\theHsection}{appendix.\Alph{section}}
\renewcommand{\theHsubsection}{appendix.\Alph{section}.\arabic{subsection}}
\renewcommand{\theHequation}{appendix.\Alph{section}.\arabic{equation}}
\section{Implementation Details}
Detailed architectural specifications and training hyperparameters are summarized in Table \ref{tab:arch_details} and Table \ref{tab:train_details}, respectively. Regarding the optimization objectives, the pixel-level MSE loss is computed over the entire image to ensure global reconstruction fidelity, whereas the auxiliary semantic loss is restricted to the masked patches to enforce the model's ability to infer missing structural context. For the adversarial refinement stage, we adopt the GAN loss formulation and discriminator architecture following RAE \cite{RAE}, with the distinction that our decoder and discriminator are optimized simultaneously from the beginning of the fine-tuning phase. During inference, we use the Classifier-Free Guidance (CFG) strategy designed by TARFlow \cite{zhai2024tarflow} exclusively within the final deep block. We strictly follow the setup of ADM~\cite{dhariwal2021diffusion} for data augmentation and evaluation.
\begin{table}[h]
\centering
\caption{\textbf{Detailed Architecture Configurations of MIMFlow-L.} $K$ denotes the number of latent tokens, and $D$ denotes the latent dimensionality.}
\label{tab:arch_details}
\begin{tabular}{@{}llc@{}}
\toprule
\textbf{Module} & \textbf{Hyperparameter} & \textbf{Value} \\ \midrule
\multirow{5}{*}{Masked Encoder $E_\phi$} & Backbone & ViT-B \\
 & Patch Size & $16 \times 16$ \\
 & Layers / Hidden Dim / Heads & 12 / 768 / 12 \\
 & Input Resolution & $256 \times 256$ \\
 & Masking Ratio & 0.4 -- 0.6 \\ \midrule
\multirow{4}{*}{Latent Space} & Number of Latent Tokens ($K$) & 128 \\
 & Latent Dimension ($D$) & 64 \\
 & Positional Encoding & 1D Learnable \\
 & Latent Noise Scale ($\sigma$) & 0.3 \\ \midrule
\multirow{3}{*}{Normalizing Flow $f_\theta$} & Architecture & Improved-STARFlow-L \\
 & Blocks / Layers & 8 / 2$\times$7+20 \\
 & Hidden Dimension & 1024 \\ \midrule
\multirow{4}{*}{Generative Decoder $D_\psi$} & Backbone & ViT-B \\
 & Layers / Hidden Dim / Heads & 12 / 768 / 12 \\
 & Output Resolution & $256 \times 256$ \\ \bottomrule
\end{tabular}
\end{table}

\begin{table}[t]
\centering
\caption{\textbf{Training Hyperparameters.} Phase 1 is the joint optimization of VAE and NF; Phase 2 is the adversarial refinement of the decoder.}
\label{tab:train_details}
\begin{tabular}{@{}lcc@{}}
\toprule
\textbf{Hyperparameter} & \textbf{Phase 1: Joint Training} & \textbf{Phase 2: Decoder Fine-tuning} \\ \midrule
Total Epochs & 90 & 2 \\
Batch Size & 256 & 256 \\
Optimizer & AdamW & AdamW \\
EMA update & 0.9999 & 0.9999 \\
Learning Rate & $1 \times 10^{-4}$ & $1 \times 10^{-4}$ \\
LR Schedule & Constant & Constant \\
Weight Decay & $1 \times 10^{-4}$ & $1 \times 10^{-4}$ \\
Adam $(\beta_1, \beta_2)$ & (0.9, 0.95) & (0.9, 0.95) \\ \midrule
\textbf{Loss Weights} & & \\
Flow Loss ($\beta$) & 1.0 & -- \\
Reconstruction ($\ell_2$) & 1.0 & 1.0 \\
Perceptual (LPIPS) & 1.1 & 1.1 \\
GAN Loss ($\alpha$) & -- & 0.05 \\ \midrule
\textbf{Hardware} & \multicolumn{2}{c}{8 $\times$ NVIDIA A100 (80GB)} \\ \bottomrule
\end{tabular}
\end{table}

\section{Spectral Analysis of the Jacobian}

To characterize the geometric properties and the stability of the Normalizing Flow (NF) mapping $f: \mathcal{X} \to \mathcal{Z}$, we analyze the spectral properties of the Jacobian matrix $J(\mathbf{x}) = \nabla_{\mathbf{x}} f(\mathbf{x})$ in ~\cref{sec:analysis}. Due to the high dimensionality of the data space, explicitly computing or storing the $D \times D$ Jacobian matrix is computationally intractable. Instead, we employ \textit{matrix-free iterative methods} based on Automatic Differentiation (AD) primitives to estimate the extreme singular values.

\subsection{Estimation of the Maximum Singular Value ($\sigma_{\max}$)}
The spectral norm, or the maximum singular value $\sigma_{\max}(J)$, is computed using the \textbf{Power Iteration} method applied to the symmetric positive semi-definite operator $J^T J$. The algorithm avoids explicit matrix construction by leveraging \textbf{Jacobian-Vector Products (JVP)} and \textbf{Vector-Jacobian Products (VJP)}:
\begin{equation}
    \mathbf{v}_{k+1} = \frac{J^T (J \mathbf{v}_k)}{\|J^T (J \mathbf{v}_k)\|},
\end{equation}
where $J \mathbf{v}_k$ is computed via a JVP, and the subsequent multiplication by $J^T$ is performed via a VJP. Upon convergence, the maximum singular value is obtained as $\sigma_{\max}(J) = \|J \mathbf{v}\|$.

\subsection{Estimation of the Minimum Singular Value ($\sigma_{\min}$)}
To estimate the minimum singular value, we implement the \textbf{Inexact Shifted Inverse Iteration}. We consider the regularized operator $A = J^T J + \alpha I$, where $\alpha > 0$ is a small numerical shift (Tikhonov regularization) introduced to ensure strict positive definiteness and numerical stability. In each iteration, we solve the linear system:
\begin{equation}
    (J^T J + \alpha I) \mathbf{w} = \mathbf{v}_k,
\end{equation}
for $\mathbf{w}$ using the \textbf{Conjugate Gradient (CG)} algorithm. This approach is ``matrix-free'' as CG only requires the evaluation of the operator-vector product $A\mathbf{v}$, which is efficiently computed using the JVP-VJP sequence. The smallest singular value is then recovered from the converged eigenvalue $\lambda_{\min}$ of $A$:
\begin{equation}
    \sigma_{\min}(J) = \sqrt{\max(\lambda_{\min} - \alpha, 0)}.
\end{equation}
The local conditioning of the NF is subsequently evaluated via the condition number $\kappa(J) = \sigma_{\max} / \sigma_{\min}$.

\section{Extended Experimental Results}

\begin{center}
\centering
\small
\setlength{\tabcolsep}{3pt}
\captionsetup{hypcap=false}
\captionof{table}{\textbf{Progressive ablation} on ImageNet $256 \times 256$ with 10K samples, evaluated without CFG.}
\label{tab:progressive}
\resizebox{\textwidth}{!}{
\begin{tabular}{lccccc}
\toprule
Configuration & rFID$\downarrow$ & gFID$\downarrow$ & IS$\uparrow$ & Prec.$\uparrow$ & Rec.$\uparrow$ \\
\midrule
Baseline (e2e, SD-VAE, 256tok, GAN) & 2.74 & 11.24 & 86.0 & 0.68 & 0.63 \\
+ Learnable Tokens, \:- GAN Loss & 4.15 & 19.52 & 63.1 & 0.52 & 0.66 \\
\quad + Aux Loss (DINO+CLIP) & 14.75 & 18.71 & 91.0 & 0.65 & 0.62 \\
\quad\quad \textbf{+ Masking (0.4--0.6)} & \textbf{3.81} & \textbf{10.14} & \textbf{105.2} & \textbf{0.64} & \textbf{0.66} \\
\quad\quad\quad + GAN FT (Full MIMFlow) & 1.47 & 6.15 & 130.0 & 0.77 & 0.71 \\
\bottomrule
\end{tabular}
}
\end{center}

\subsection{Progressive ablation.}
As shown in \cref{tab:progressive}, this ablation progressively isolates the contribution of each component under a 10K-sample evaluation without CFG. Starting from the end-to-end SD-VAE baseline with 256 tokens and GAN training, replacing the latent interface with learnable tokens while removing the GAN loss degrades gFID from 11.24 to 19.52, indicating that learnable tokens alone do not explain the final gain. Adding DINO+CLIP auxiliary supervision improves semantic confidence as reflected by IS, but only marginally changes gFID (19.52$\to$18.71). The decisive improvement appears when the 0.4--0.6 masking bottleneck is introduced, reducing gFID by 46\% (18.71$\to$10.14) and surpassing the GAN-trained baseline. This supports our claim that masking provides the primary structural constraint for simplifying the latent distribution modeled by the NF, rather than the gain coming mainly from auxiliary supervision. The final GAN fine-tuning stage further recovers low-level fidelity and texture detail, improving rFID from 3.81 to 1.47 and gFID from 10.14 to 6.15.

\begin{table}[t!]
    \centering
    \small
    \renewcommand{\arraystretch}{1.15}
    \begin{minipage}[t]{0.53\textwidth}
        \centering
        \setlength{\tabcolsep}{3.2pt}
        \caption{\textbf{Ablation of MIM weight}}
        \label{tab:mim}
        \begin{tabular}{lccccc}
            \toprule
            MIM & rFID$\downarrow$ & gFID$\downarrow$ & IS$\uparrow$ & Prec.$\uparrow$ & Rec.$\uparrow$ \\
            \midrule
            \rowcolor[gray]{0.95} \textbf{1} & \textbf{3.60} & \textbf{12.46} & \textbf{92.3} & \textbf{0.70} & \textbf{0.66} \\
            5 & 4.10 & 13.33 & 85.6 & 0.69 & 0.64 \\
            10 & 13.73 & 26.19 & 61.8 & 0.46 & 0.55 \\
            \bottomrule
        \end{tabular}
    \end{minipage}
    \hfill
    \begin{minipage}[t]{0.45\textwidth}
        \centering
        \setlength{\tabcolsep}{4.5pt}
        \caption{\textbf{Efficiency Analysis}}
        \label{tab:performance_metrics}
        \begin{tabular}{ccc}
            \toprule
            Token & Train(ms/iter) & Inference(s) \\
            \midrule
            \rowcolor[gray]{0.95}\textbf{128} & \textbf{183} & \textbf{1.60} \\
            256 & 352 & 3.64 \\
            1024 & 1515 & 25.3 \\
            \bottomrule
        \end{tabular}
    \end{minipage}

\end{table}

\subsection{Ablation on Masked Reconstruction Weight}
In standard Masked Image Modeling (MIM), such as MAE~\cite{MAE}, the reconstruction loss is typically computed only on the masked patches. However, for generative modeling, maintaining global pixel-level fidelity is crucial for high-quality synthesis. To balance these objectives, we investigate the impact of the MIM weight, which scales the reconstruction loss of the masked regions relative to the unmasked ones (the latter fixed at a weight of 1).

As shown in \cref{tab:mim}, the optimal performance is achieved when the MIM weight is set to 1, effectively treating masked and unmasked regions with equal importance during joint optimization. Increasing the weight to 5 or 10 leads to a noticeable degradation in both reconstruction (rFID) and generation (gFID) metrics. This performance drop suggests that over-weighting the masked regions may bias the model toward local patch-filling at the expense of global structural coherence, thereby distorting the learned semantic manifold.

\subsection{Efficiency Analysis}
A key advantage of our MIMFlow is its high efficiency, achieved through a significantly reduced token budget. While existing methods typically rely on 256 or even 1024 tokens to represent sequences, our approach operates effectively with only 128 tokens. As demonstrated in Table \ref{tab:performance_metrics}, we first isolate the latency effect of token count using the Improved STARFlow backbone. On an H20 GPU, the 128-token setting achieves a training latency of 183ms per iteration (batch size 32) and an inference latency of 1.60s per sample. Compared to the 1024-token setting, it provides an 8.3$\times$ speedup in training and a 15.8$\times$ acceleration in inference.

\subsection{Semantic Evolution across Flow Depth}
\begin{figure}[t!]
    \centering
    \includegraphics[width=\linewidth]{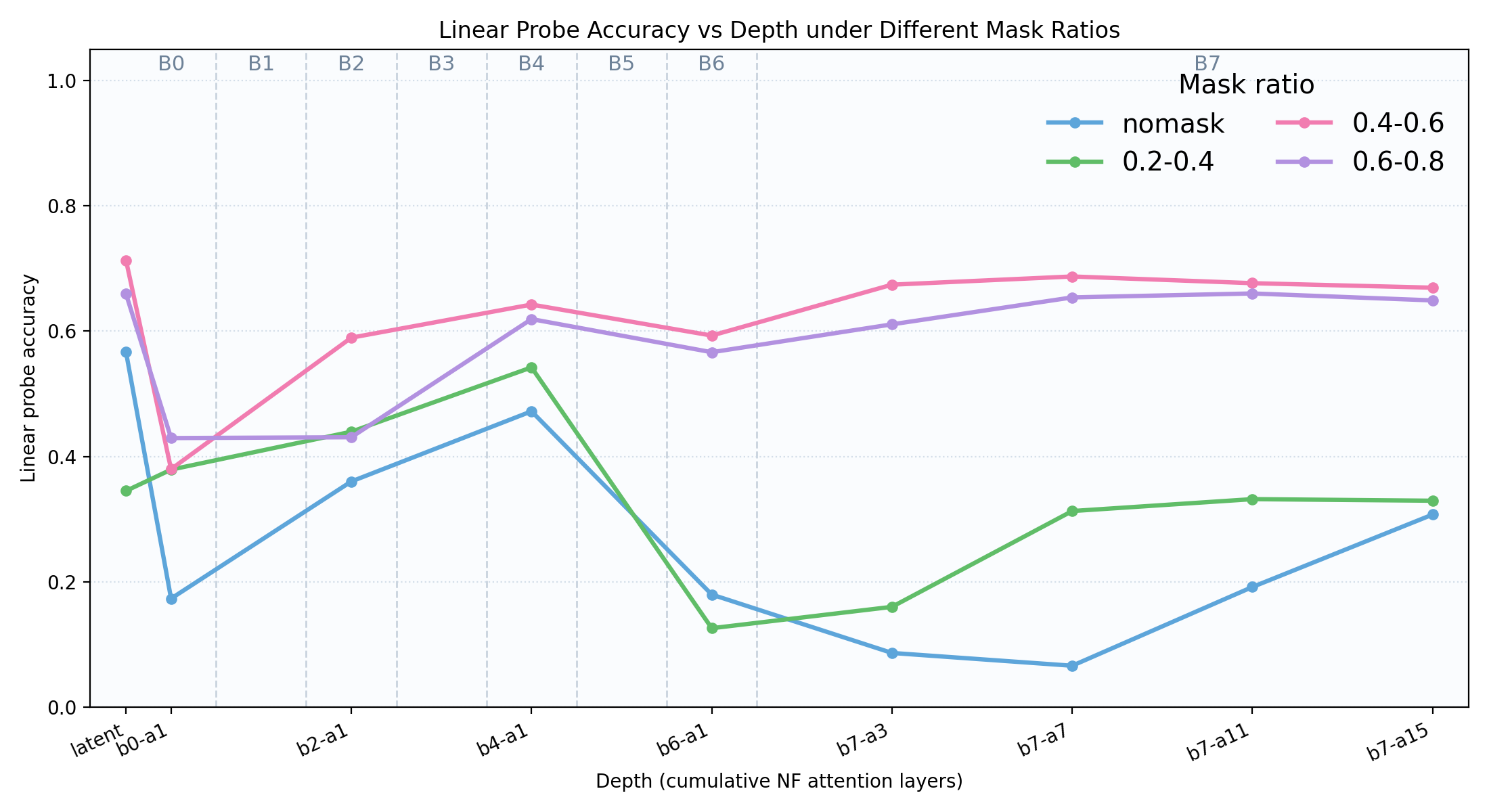}
    \captionsetup{hypcap=false}
    \captionof{figure}{\textbf{Linear Probe Accuracy vs Depth} under Different Mask Ratios.}
    \label{fig:linearprobe}
\end{figure}

To investigate the feature abstraction capabilities of Normalizing Flows, we conduct linear probing on the intermediate representations at various depths of the NF. As illustrated in \cref{fig:linearprobe}, while the encoder's latent space exhibits high classification accuracy (facilitated by the MIM objective), this accuracy does not increase—and occasionally plateaus or slightly declines—as the depth of the NF increases.

This observation reveals a fundamental characteristic of Normalizing Flows: while they excel at complex distribution warping via bijective mappings, they lack the inherent ability to perform further hierarchical feature abstraction or semantic distillation. This finding underscores the necessity of our MIMFlow paradigm, which delegates the burden of semantic modeling to the encoder through Masked Image Modeling, allowing the subsequent flow blocks to focus purely on probabilistic density estimation within a well-structured manifold.

\bibliographystyle{splncs04}
\bibliography{main}
\end{document}